# Improvement and Enhancement of YOLOv5 Small Target Recognition Based on Multi-module Optimization

Qingyang Li [1,a*]
[1]Xi'an Jiaotong University
Faculty of Electronic and Information Engineering
Xi'an, China
Corresponding author:
[a]likon2101@ieee.org

Yuchen Li [1,b]
[1]Xi'an Jiaotong University
Faculty of Electronic and Information Engineering
Xi'an, China
[b]yuchenli@ieee.org

Hongyi Duan [1,c]
[1]Xi'an Jiaotong University
Faculty of Electronic and Information Engineering
Xi'an, China
[c]dann_hiroaki@ieee.org

JiaLiang Kang [2,d]
[1]Xi'an Jiaotong University
Faculty of Electronic and Information Engineering
Xi'an, China
[d]2216113084@stu.xjtu.edu.cn

Jianan Zhang [2,e]
[2]Shanghai University of Finance and Economics University
School of Mathematica
Shanghai, China
[e]zjaqifei@ieee.org

Xueqian Gan [3,f]
[3]Tianjin University
School of Science
Tianjin, China
[f]cruise@tju.edu.cn

Ruotong Xu [3,g]
[1]Xi'an Jiaotong University
School of Management
Xi'an, China
[g]ruotongxu@ieee.org

*Abstract*: In this paper, the limitations of YOLOv5s model on small target detection task are deeply studied and improved. The performance of the model is successfully enhanced by introducing GhostNet-based convolutional module, RepGFPN-based Neck module optimization, CA and Transformer's attention mechanism, and loss function improvement using NWD. The experimental results validate the positive impact of these improvement strategies on model precision, recall and mAP. In particular, the improved model shows significant superiority in dealing with complex backgrounds and tiny targets in real-world application tests. This study provides an effective optimization strategy for the YOLOv5s model on small target detection, and lays a solid foundation for future related research and applications.

## I. Introduction and Review

In today's industrial production and daily life, small target recognition plays an increasingly important role. In industrial production, whether it is the detection of minute defects or the precise positioning of parts on a high-speed production line, the accurate identification of small targets is the key to improving production efficiency and product quality. For example, in semiconductor manufacturing, early detection of defects at the micron or even nanometer level can significantly reduce production costs. In everyday life, small target recognition is used everywhere, from cell detection in medical images to tracking small drones in security surveillance. Research on small target recognition has attracted a great deal of academic attention. Traditional approaches are mainly based on feature engineering and hand-designed algorithms[1-2]. However, with the rise of deep learning, Convolutional Neural Network (CNN) based methods are starting to dominate[3-4]. The YOLO (You Only Look Once) series is one of the best of the best, known for its excellent speed and accuracy. Especially YOLOv5, this version further optimizes the performance and speed[5-6]. Nevertheless, YOLOv5 still has some limitations in small target recognition. Many researchers have tried to improve the performance of small target recognition by combining techniques such as feature pyramid, data augmentation and attention mechanism[6-8]. Despite some progress in research, there are still many challenges in small target recognition, especially in complex backgrounds and different environmental conditions. First, the small targets occupy a small space in the image, resulting in a feature representation that may be interfered by a large amount of background noise[9-11]. Second, YOLOv5 and its predecessors mainly focus on the detection of large targets, while the detection performance for small targets is not satisfactory[12-13]. In addition, the existing loss functions may not be sufficient to handle the subtle differences between small targets and background[14-15].

To address the above problems, this paper proposes a YOLOv5 small target recognition method based on multi-module optimization. Specifically, we introduce GhostNet in the convolutional module to enhance the feature extraction ability of the model. In the Neck part of the network, we employ RepGFPN to better fuse multi-scale features. In addition, we combine CA and Transformer to enhance the attention mechanism of the model. Finally, we introduce the Normalized Gaussian Wasserstein Distance loss as a loss function to better distinguish the small targets from the background. With these improvements, our method significantly improves the performance of small target recognition on multiple datasets.

## II. YOLOV5 OVERVIEW AND ITS BENEFITS

YOLO, which stands for "You Only Look Once", is a revolution in target detection. Since its debut in 2016, it has gone through several iterations, each bringing new innovations.YOLOv5 is the latest product of this series, representing the current research frontier, and as the latest version of this series, it continues to follow its overall design concept. In this paper, we use one of them, i.e., Yolov5s, whose network structure is shown in Figure 1.

First, it treats the entire image as a large scene and parses it all at once. This means that it does not need to scan the entire image multiple times, thus greatly improving the detection speed compared to traditional sliding window and region proposal methods. Specifically, it divides the image into an SxS grid, each of which is responsible for predicting the objects within its boundaries. Each grid predicts B bounding boxes and the class probabilities associated with these bounding boxes.

The prediction of bounding boxes is based on anchor frames which are predefined and represent different shapes and sizes. The model then learns how to adjust these anchor frames to the image content during training to more accurately match objects in the image.YOLOv5 also introduces multi-scale feature extraction, which allows it to detect objects of different sizes simultaneously. Specifically, it performs detection on multiple layers of feature maps so that it can capture global

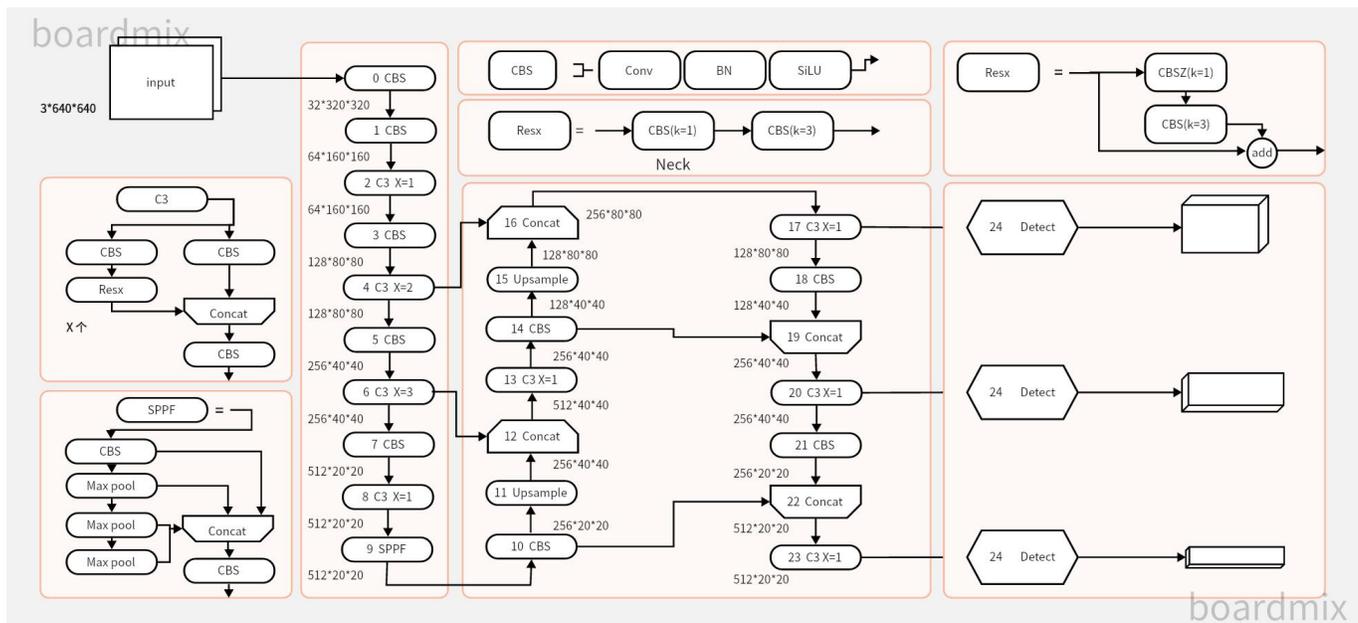

Figure 1:The complete structure of yoloV5s

information about large objects as well as detailed features of small objects.

The advantages and innovations of YOLOv5 can be summarized as follows:

- Lightweight design: although YOLOv5 maintains a high accuracy rate, its model size is relatively small. This is due to its optimized network structure and parameters. This makes YOLOv5 not only suitable for high-performance computing platforms, but also ideal for deployment on resource-constrained devices such as mobile devices and embedded systems.

- Improved anchor frame strategy: YOLOv5 further optimizes its anchor frame strategy. It uses more anchor frames and these anchor frames better match the real object shapes in the training data. This allows the model to match objects more accurately during prediction, thus improving detection performance.

- Enhanced Data Enhancement: data enhancement is the key to improving the generalization ability of the model. yolov5 introduces a series of new data enhancement techniques, including random cropping, rotation, color warping, and so on. These techniques greatly increase the diversity of training data, enabling the model to better handle a wide variety of scenes and conditions.

- Further optimization of the model structure: the network structure of YOLOv5 is optimized at several levels. For example, it introduces new activation functions and regularization strategies to reduce overfitting and improve the robustness of the model. In addition, it optimizes the feature extraction and fusion mechanisms so that the model can better capture the detailed information in the image.

### III. MULTI-MODULE IMPROVEMENTS FOR YOLOv5s

#### A. Convolutional Module Improvement Based on GhostNet Networks

In the original YOLOv5, the convolution module consists of CONV+BN (batch normalization) + SiLU (sigmoid linear unit). Although this structure achieves good performance in most of the scenes, it has some limitations in recognizing small targets. In particular, when small targets are mixed with complex backgrounds in an image, the regular convolution module may ignore them or have difficulty in localizing them accurately due to the small space they occupy in the image. In addition, regular convolution may introduce more background noise that interferes with the model's recognition of small targets.

GhostNet, a new lightweight convolutional neural network, aims to reduce the computational burden by generating "ghost" features while maintaining high performance. The core idea is that instead of computing all the feature maps directly, GhostNet generates additional feature maps from a smaller set of features by linear transformation.

The Ghost Bottleneck is a key component of GhostNet, and the following Figure 2. illustrates the Ghost Bottleneck module.

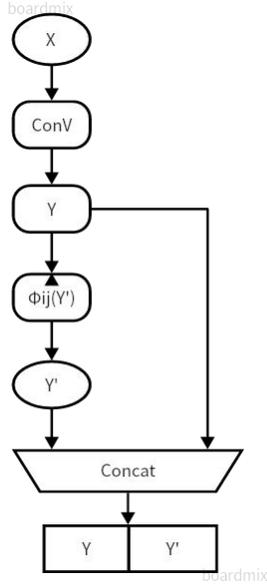

Figure 2. Ghost Bottleneck Module Schematic

First for a given graph $X \in R^{c \times h \times w}$ the convolution is computed by convolving $Y = X * f + b$ to get $Y' = X * f'$ where $f' \in R^{c \times k \times k \times w}$ is the convolution kernel for the layer. Then more feature maps are obtained using the following linear variation

$$Y_{ij} = \Phi_{i,j}(Y_i^{'}), \forall i = 1, \cdots, m; j = 1, \cdots, s$$

where $Y_i$ is the ith feature map and $\Phi_{i,j}$ is the linear transformation function of the jth feature transformation for the i feature map

The key advantages of GhostNet for small target recognition are focused on:

- Higher feature resolution: due to the design of GhostNet, it is able to generate more feature maps without increasing the amount of computation. This means that the network can better capture the detailed information in the image, which helps to improve the detection accuracy of small targets.

- Reducing background noise: the linear transformation strategy of GhostNet can better filter out the background noise and reduce the interference of small target recognition.

- Lightweight design: GhostNet has a small model size and computational complexity, which means it can run on resource-constrained devices and enable real-time small target detection.

- Stronger robustness: due to the design and structure of GhostNet, it is more robust to various changes and distortions in images, which helps to improve the performance of small-target detection in various scenes and conditions.

#### B. RepGFPN-based optimization of Neck modules

In the YOLOv5s design, the Neck part is clearly influenced by the PANet structure. By adding bottom-to-top paths in the FPN-based design, PANet successfully enhances the model's ability to capture features. However, for the recognition of small targets, there is still room for improvement in the performance of YOLOv5s. This is mainly due to the fact that small targets account for a small percentage of the whole image, and the large downsampling strategy of YOLOv5s may lead to insufficient feature capture of these targets. This results in a lack of feature information for small targets in the deeper network of the model, while in the shallow part, these features are underutilized.

Based on the above shortcomings, we choose RepGFPN to optimally reconstruct the Neck part of YOLOv5s, the structure of which is illustrated in Figure. 3. The core strategy is to fuse multi-scale feature maps using up and down sampling, and to further enhance the complementarity of these features through a specially designed Fusion module. The implementation of Fusion traditionally relies on the C3 module, and we modify C3 and integrate the concepts of ELAN and Rep to obtain better fusion features. The structure of the improved Fusion module is shown in Figure 4.

The Fusion Convolution (FC) module revolutionizes the design ideas of RepConvN and CBS. Compared to the traditional C3 module, this new FC module not only enhances its feature extraction and fusion capabilities, but also stabilizes the speed of detection by utilizing a multi-branch design. We choose this combination of RepConvN and CBS as the core of FC mainly based on the fact that it can ensure the accuracy of detection while reducing the model complexity.

In summary, by introducing the RepGFPN strategy to optimize the Neck part, our model was able to integrate features of different layers more accurately. This optimization strategy not only strengthens the synergistic effect among features, but also significantly improves the recognition accuracy of small targets in a variable scene.

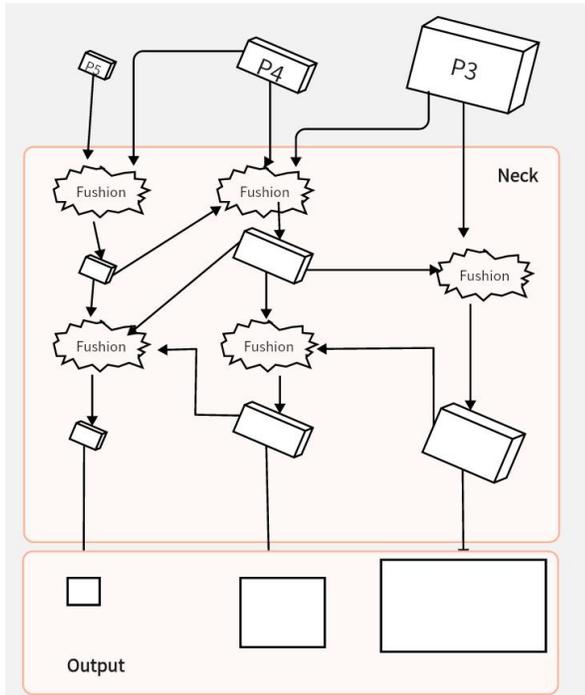

Figure 3. RepGFPN module structure

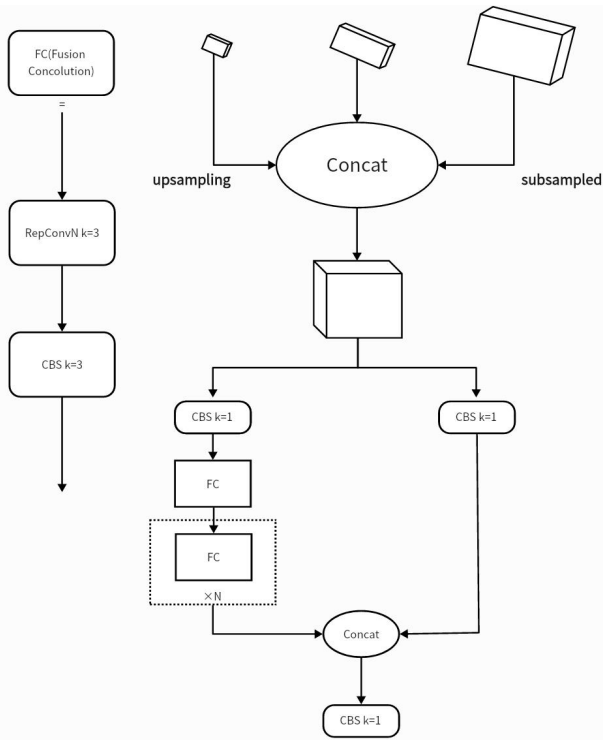

Figure 4. Improved Fusion module structure

## C. The addition of CA+Transformer based Attention Mechanism

Attention mechanisms play a key role in deep learning models, especially in complex tasks such as small target detection. Introducing an attention mechanism can improve the performance of the model by making it more focused on key parts of the image. This is especially important in small target detection tasks, as small targets often occupy a small space in the image and the model needs to be able to recognize and focus on these tiny targets. To better achieve this goal, two attention mechanisms are chosen in this paper: the CA (Coordinate Attention) and the Transformer.

### 1) Coordinate Attention (CA) module

The Coordinate Attention (CA) module is an innovative attention mechanism introduced in recent years for visual tasks. Unlike traditional attention mechanisms that focus on the content of features, the CA module pays more attention to the position or coordinate information on the feature map. This heightened attention to location allows the model to better capture local contextual information when processing images.

In a specific operation, the CA module first calculates the distribution of coordinate attention in the horizontal (i.e., row) and vertical (i.e., column) directions of the feature map, respectively. This process is obtained by performing a summation operation on the feature map and normalizing it. After obtaining the attention distributions in these two directions, the CA module applies them back to the original feature map to form a weighted feature map, which is more spatially concentrated and more capable of highlighting those regions that are beneficial to the task. The structure is shown in the following Figure5

For small target detection, this high attention to location is especially important. Because in real images, small targets are often intertwined with the background and other objects, and their pixel occupancy in the image is very small. Traditional convolutional neural networks, when dealing with this situation, may ignore small targets because of their small pixel share, leading to an increase in false negatives in detection.The CA module, on the other hand, by reinforcing the weights of the locations of small targets, makes the model pay more attention to dealing with these locations, and thus it is more likely to detect these small targets correctly.

In addition, the computational process of the CA module is relatively simple and does not introduce much computational complexity, which makes it easy to combine with other modules to provide better performance. In summary, the introduction of the CA module is undoubtedly a wise choice for the small target detection task. It not only improves the accuracy of the model, but also performs well in terms of computational efficiency.

### 2) Transformer Encoder Module

The Transformer Encoder module, derived from the original Transformer model, has become a star structure in the field of natural language processing in recent years as shown below in the Figure 6.

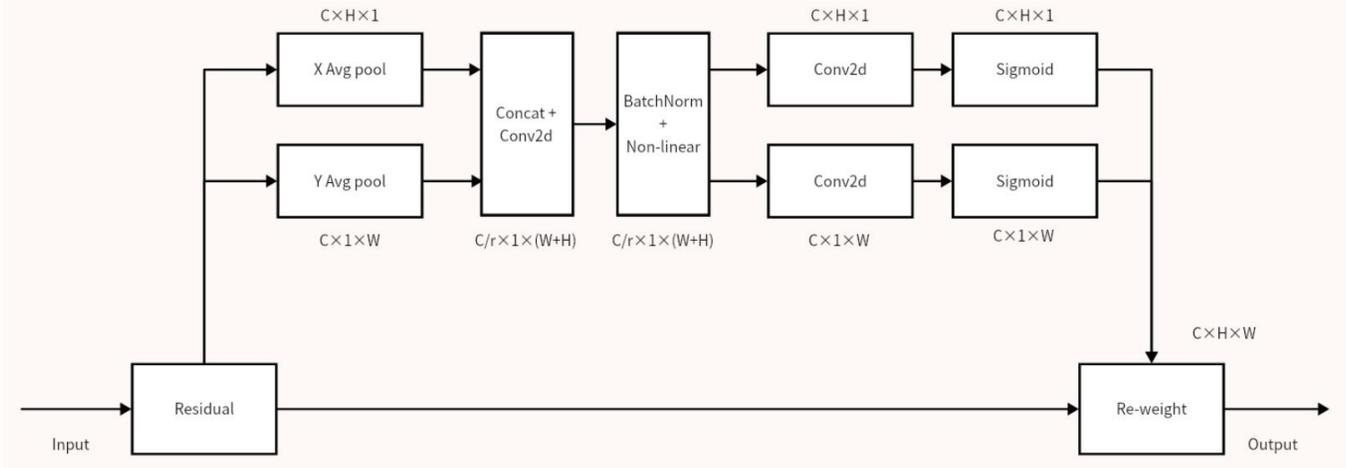
Figure 5. Coordinate Attention (CA) module structure

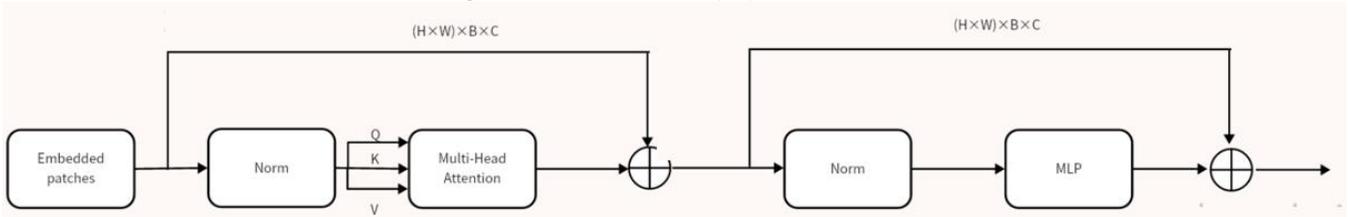
Figure 6. Transformer Encoder module structure

It is characterized by its ability to capture long-range dependencies in the input data and allows data to flow freely between different time steps. Its core component, the self-attention mechanism, emphasizes the assignment of weights to the input features, allowing the model to be more focused on the critical parts of the input. In this paper, we introduce the Transformer Encoder module in front of the head detection network. Transformer Encoder has the following features and advantages:

- Global Information Acquisition: the Transformer Encoder captures global information in the image, which allows the model to not only focus on local features, but also understand the context of the entire image, leading to more accurate detection of small targets.

- Modal Fusion Capability: Transformer Encoder is inherently good at modal fusion. This means that when data comes from different sources (e.g., different sensors or viewpoints), Transformer Encoder can effectively integrate this information to improve the detection performance of the model.

- Multi Head Attention Layer (MHA): MHA is another major feature of Transformer Encoder. It can map the captured global information to multiple different spaces, thus describing the contextual information more richly. This is especially important in small target detection as it allows the model to understand the image from multiple perspectives and thus localize small targets more accurately.

- MLP layer: in addition to MHA, Transformer Encoder contains an MLP layer. The presence of this layer allows for better convergence and prevents degradation of the network, thus enhancing the expressive power of the model.

The introduction of the Transformer Encoder module is particularly crucial in image detection tasks, especially in small target recognition. Small targets are often hidden in complex backgrounds or obscured by large targets in images. At this time, it is difficult to accurately detect small targets only relying on local information. The Transformer Encoder, with its self-attention mechanism, can capture the global information in the image. This means that even if a small target is occluded or mixed with the background in a certain part of the image, the model can recognize the small target by analyzing the whole image.

### D. NWD-based loss function improvement

The bounding box loss function used in the original YOLOv5s model is CIoU (compatible intersection over union). The CIoU loss function combines the IoU and the distance factor, aiming to solve the instability that the traditional IoU loss function may cause in bounding box regression. Although CIoU loss performs well in many target detection tasks, it still has some drawbacks in small object recognition. Small objects have a limited number of pixels, and even small localization errors may lead to significant changes in IoU, which can affect the performance of the model. In addition, the CIoU loss function may not adequately take into account the uniqueness of small object features, which may lead to unsatisfactory performance of the model on small objects.

To overcome these problems, this paper proposes to use Normalized Gaussian Wasserstein Distance (NWD) as the loss function. The NWD loss function is mathematically defined as

the Wasserstein distance between two normal distributions, which is given in the following equation:

$$NWD(N_a, N_b) = e^{(-\frac{\sqrt{W_2^2(N_a, N_b)}}{C})}$$

wher $N_a, N_b$ is NWD is the number of true frames in the neighborhood of each prediction frame and the coefficient $W_2^2(N_a, N_b)$ is calculated as:

$$W_2^2(N_a, N_b) = \left\| \left[cx_a, cy_a, \frac{w_a}{2}, \frac{h_a}{2}\right]^T, \left[cx_b, cy_b, \frac{w_b}{2}, \frac{h_b}{2}\right]^T \right\|$$

represent the center point coordinates and width and height of the two frames, respectively, and measure the relative importance between the true frame loss and the predicted frame loss.

model provides a powerful tool to measure the difference between two distributions in the feature space. Compared to CIoU, the NWD loss function has the following advantages for small object recognition:

- Stability: since the NWD loss function directly takes into account the differences between the distributions, it is more robust to the localization errors of small objects, thus improving the stability of the model.

- Feature sensitivity: the NWD loss function can better capture the unique features of small objects, thus improving the model's recognition accuracy on small objects.

- Global Consideration: the NWD loss function takes into account the global information of all the objects in the image, which allows the model to better understand the relationship between small objects and other objects in the image.

IV. EXPERIMENTAL ENVIRONMENT AND EVALUATION INDICATORS

A. Experimental environment

In this paper, we use a deep learning framework based on PyTorch, the operating system is Ubuntu 20.04 built on VMWare Workstation Pro, and the hardware is M2 Max. the key parameters are set as follows: input pixels 960x960, batch size=12, epoch=350. the dataset is collected as well as AI generated by us. A total of 1000 flat panels with imperfections (mostly cell phone as well as tablet backshells) were obtained with tiny scratches and imperfections in the panels acting as so-called mini-targets and divided into training, validation, and test sets in the ratio of 8:1:1. For the training dataset we also applied additional data enhancement techniques.

B. Evaluation indicators

In this paper, precision rate P, recall rate R and mean average precision mean mAP are selected as evaluation metrics, and the formula is:

$$P = \frac{TP}{TP + FP} = \frac{true\ postives}{true\ postives + false\ postives}$$

$$R = \frac{TP}{TP + FN} = \frac{true\ postives}{true\ postives + false\ negative}$$

$$mPA = \sum_{i=1}^{n} \frac{\int_0^1 P_i dR}{n}$$

V. RESULTS

A. Average Accuracy Comparison

As shown in the Figure7. , the training results of the improved YoloV5s and the original YoloV5s in the same situation, it can be seen that there is a considerable improvement in its accuracy after the improvement. However, both of them remain basically the same in the speed of convergence.

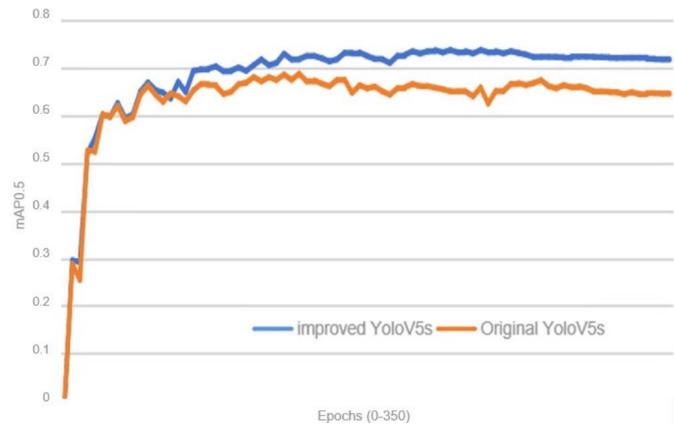

Figure 7. Comparative Results

After the text edit has been completed, the paper is ready for the template. Duplicate the template file by using the Save As command, and use the naming convention prescribed by your conference for the name of your paper. In this newly created file, highlight all of the contents and import your prepared text file. You are now ready to style your paper; use the scroll down window on the left of the MS Word Formatting toolbar.

B. Ablation experiment

Further comparisons of ablation experiments are as follows shown in the Chart1. Improvements 1, 2, 3, and 4 correspond to the improvement of the convolution module, the improvement of Neck, the addition of attention mechanism, and the improvement of loss function. And 0/1 means with/without such optimization, respectively)

TABLE I. RESULTS OF B.ABLATION EXPERIMENT

| I | II | III | IV | $mPA_{0.5}$ /% | P/% | R/% |
|---|----|----|----|----------------|------|------|
| 0 | 0  | 0  | 0  | 59.3556        | 61.3264 | 57.2305 |
| 1 | 0 | 0 | 0 | 60.4172 | 62.7481 | 58.6415 |
| 0 | 1 | 0 | 0 | 61.4166 | 62.9646 | 60.112 |
| 0 | 0 | 1 | 0 | 60.6631 | 63.7389 | 58.9305 |
| 0 | 0 | 0 | 1 | 62.3257 | 64.3861 | 60.0185 |
| 1 | 1 | 0 | 0 | 67.0323 | 70.3263 | 67.8385 |
| 1 | 0 | 1 | 0 | 67.3830 | 73.0716 | 69.9893 |
| 1 | 0 | 0 | 1 | 70.4524 | 72.8371 | 68.7314 |
| 0 | 1 | 1 | 0 | 70.3086 | 71.9829 | 69.4025 |
| 0 | 1 | 0 | 1 | 69.6061 | 73.9710 | 70.5075 |
| 0 | 0 | 1 | 1 | 69.5795 | 74.9883 | 70.3885 |
| 1 | 1 | 1 | 0 | 76.3291 | 77.3281 | 72.9215 |
| 1 | 1 | 0 | 1 | 76.7798 | 75.7177 | 73.1085 |
| 1 | 0 | 1 | 1 | 74.8264 | 75.0694 | 73.1236 |
| 0 | 1 | 1 | 1 | 74.1061 | 75.5282 | 71.6047 |
| 1 | 1 | 1 | 1 | 79.3174 | 80.4066 | 76.6193 |

Before optimizing the YOLOv5 model, we first observed the benchmark performance of the model without any optimization strategy. Specifically, the model has an average precision mean of 59.355%, a precision of 61.326%, and a recall of 57.2305%. This provides a reference standard for subsequent optimization experiments.

GhostNet Improvement of Convolutional Module : GhostNet, as a lightweight and efficient convolutional module, aims at extracting richer features through less computation. Experimental results show that when GhostNet is introduced, the average accuracy of the model improves by about 1.06 percentage points on average. This strongly suggests that GhostNet can capture the detailed information of the image more deeply while maintaining the computational efficiency, especially for feature extraction of small targets.

RepGFPN Improvement for Neck: RepGFPN provides a new way of feature fusion for the model, especially when dealing with features at different scales. With the introduction of RepGFPN, the model achieves a significant improvement in average precision mean and recall. This confirms that RepGFPN has significant benefits for small target recognition as it can better fuse multi-scale features and enhance the model's sensitivity to small targets.

Attention Mechanism Introduced by CA+Transformer: the attention mechanism is designed to allow the model to focus on key parts of the image. Although CA+Transformer improves the accuracy of the model, its impact on the average precision mean and recall is relatively limited. This may indicate that the attention mechanism is more about helping the model to localize the target rather than just recognizing it. This is especially critical for small targets, which make up a small percentage of the image and are easily overlooked.

Normalized Gaussian Wasserstein Distance Improvement of Loss Function: the recall of the model is significantly improved after the introduction of the new loss function. This shows that the Normalized Gaussian Wasserstein Distance loss is able to handle positive and negative samples in a more balanced way, making the model more robust in detecting small targets.

In summary, when these four optimization strategies are used jointly, the model performance is significantly improved, with the highest average precision mean of 79.317%, and precision and recall of 80.406% and 76.619%, respectively. This not only confirms the effectiveness of each individual optimization strategy, but also shows the improvement in model performance when they work together.

### C. Comparison of Selected Test Sets

In order to further verify the performance of the improved Yolov5, we input a considerable portion of the test set into the model before and after the improvement for comparative testing, and in order to visualize the advantages and disadvantages of the model, we selected three representative pictures for illustration(as shown in the Figure8. Figure9. and Figure10. ). On each left is the original yolov5s, and on the right is the improved yolov5s.

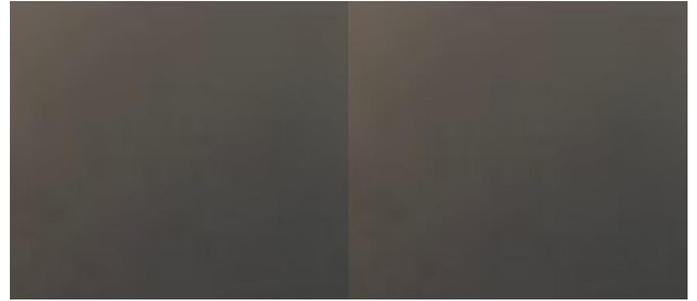

Figure 8. Representative example1

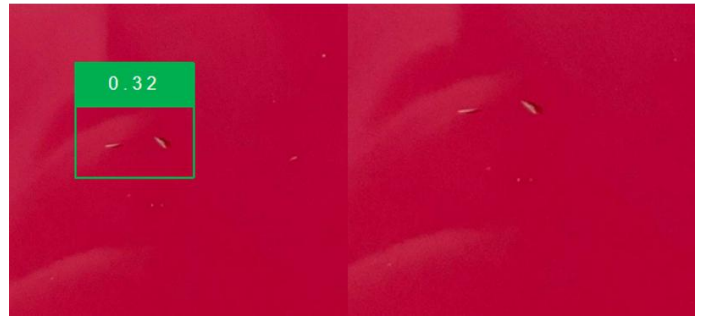

Figure 9. Representative example2

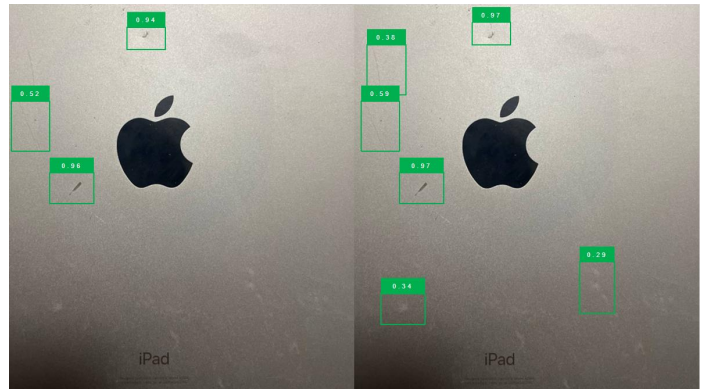

Figure 10. Representative example3

All three images were taken from the flat backside of a branded device that may have had imperfections. For the board surface that does not have any scratches itself, both the improved and the new yolov5 algorithms show good performance and do not recognize any targets. However, for the board shown in the second image, the original yolov5s algorithm mistakenly recognizes the presence of foreign objects on the surface as scratches or blemishes, while the improved algorithm avoids this error perfectly. The third image shows a board full of blemishes and scratches, and the improved algorithm outperforms the original algorithm both in terms of the number and accuracy of recognitions.

In summary, the improved YOLOv5s model can recognize scratches and defects on the surface of the flat panel more accurately than the original YOLOv5s model. This also proves that the improved YOLOv5s has superior performance in recognizing small targets.

## VI. Summary and outlook

The core objective of this study is to optimize the YOLOv5s model for small target detection tasks. After a series of multi-module improvements, including the GhostNet-based convolution module, the RepGFPN-based Neck module, the introduction of CA and Transformer's attention mechanism, and the adoption of NWD as a new loss function, we successfully improve the model's performance on small target detection.

Specific experimental results show that the improved models provide significant improvements in precision, recall, and mAP for small target detection compared to the original YOLOv5s model. The effectiveness of these improved strategies is also validated by ablation experiments, confirming the positive contribution of each optimization strategy to the performance.

In addition, real-world test set comparison experiments have demonstrated the superior performance of the improved model in recognizing scratches and imperfections on flat surfaces, especially its robustness and accuracy in dealing with complex backgrounds and tiny targets.

For future research, we suggest: deepening the study of new attention mechanisms, especially for the processing of spatial location information of small targets; exploring more efficient network structures, such as NAS technology, to further improve the computational efficiency of the model; expanding to more practical application scenarios, such as UAVs, medical images, etc., to validate the broad applicability of the model; incorporating more unsupervised or semi-supervised learning methods that reduce the dependence on a large amount of labeled data.

## VII. Acknowledgement

I am deeply indebted to Yuchen Li, Jialiang Kang, Jianan Zhang, Xueqian Gan, and Ruotong Xu for their invaluable contributions and unwavering support throughout the development of this paper. Their insights, expertise, and commitment were pivotal in shaping this research. Special accolades are reserved for Hongyi Duan, whose exemplary scientific acumen and technical prowess greatly elevated the quality of our work. Without the collective efforts of this outstanding team, the accomplishments documented in this paper would not have materialized. Their collaboration was the cornerstone of this research, and I am truly grateful for their dedication.